%% file: main.tex
\pdfoutput=1
\documentclass[10pt,twocolumn,letterpaper]{article}

\usepackage[pagenumbers]{wacv} % To force page numbers, e.g. for an arXiv version

\usepackage[accsupp]{axessibility}
\usepackage{amsmath}
\usepackage{amssymb}
\usepackage{booktabs}
\usepackage{pifont}
\usepackage{color}
\usepackage[table]{xcolor}
\usepackage{bm}
\usepackage{multirow}
\usepackage{soul}
\usepackage{bbm}
\usepackage{xurl}
\usepackage{graphicx}

\definecolor{citecolor}{HTML}{0071bc}
\definecolor{lightgray}{gray}{0.9}
\usepackage[pagebackref=true,breaklinks=true,colorlinks,citecolor=citecolor,bookmarks=false]{hyperref}

\usepackage[capitalize]{cleveref}
\crefname{section}{Sec.}{Secs.}
\Crefname{section}{Section}{main}
\Crefname{table}{Table}{Tables}
\crefname{table}{Tab.}{Tabs.}
\Crefname{figure}{Figure}{Figures}
\crefname{figure}{Fig.}{Figs.}

\newcommand{\bracket}[1]{\textless #1\textgreater}

\def\wacvPaperID{2066} % *** Enter the WACV Paper ID here
\def\confName{WACV}
\def\confYear{2026}

\begin{document}

%%%%%%%%% TITLE 
\title{Harnessing Object Grounding for Time-Sensitive Video Understanding}

\author{Tz-Ying Wu \quad Sharath Nittur Sridhar \quad Subarna Tripathi \\
Intel\\
{\small\tt\{tz-ying.wu, sharath.nittur.sridhar, subarna.tripathi\}@intel.com}
}
\maketitle

%%%%%%%%% ABSTRACT
\input{0_abstract.tex}

%%%%%%%%% BODY TEXT
\input{1_introduction.tex}

\input{2_related_works.tex}

\input{3_method.tex}

\input{4_experiments.tex}
\input{5_conclusion.tex}

\clearpage
%%%%%%%%% REFERENCES
{\small
% \bibliographystyle{ieee_fullname}
% \bibliography{egbib}

\input{main.bbl}
}

\end{document}

%% file: 0_abstract.tex
\begin{abstract}
We propose to improve the time-sensitive video understanding (TSV) capability of video large language models (Video-LLMs) with grounded objects (GO). We hypothesize that TSV tasks can benefit from GO within frames, which is supported by our preliminary experiments on LITA, a state-of-the-art Video-LLM for reasoning temporal localization. While augmenting prompts with textual descriptions of these object annotations improves the performance of LITA, it also introduces extra token length and susceptibility to the noise in object-level information. To address this, we propose {\bf GO-Tokenizer}\footnote{Project page: \url{https://intelailabpage.github.io/2025/12/08/go-video.html}.}, a lightweight add-on module for Video-LLMs leveraging off-the-shelf object detectors to encode compact object information on the fly. Experimental results demonstrate that pretraining with GO-Tokenizer outperforms the vanilla Video-LLM and its counterpart, utilizing textual descriptions of objects in the prompt. The gain generalizes across different models, datasets, and video understanding tasks, such as reasoning temporal localization and dense captioning.
\end{abstract}

%% file: 1_introduction.tex
\section{Introduction}
Extending the ability of Large-Language-Models (LLMs) to the video modality is an emerging field that seeks to equip Video-LLMs~\cite{lin2024videollavalearningunitedvisual, damonlpsg2023videollama,cheng2024videollama2advancingspatialtemporal, maaz2024videochatgptdetailedvideounderstanding} with the ability to understand long and complex video, and respond to user queries for various downstream tasks, such as video summarization~\cite{sharghi2017queryfocusedvideosummarizationdataset,Song_2015_CVPR} and video captioning~\cite{activityNetCaptionskrishna2017dense,Youcook2ZhXuCoAAAI18}.
To leverage the knowledge embedded in pre-trained LLMs, most Video-LLMs learn video representations that share a latent space with text, allowing features from both modalities to be jointly processed during training and inference.

While Video-LLMs inherit the knowledge and reasoning capabilities from pretrained LLMs, the limitation of LLMs in modeling long sequences remains. This is particularly worrisome since a video usually contains several frames, and the spatial resolution of each frame could be large.
Given the sequence length limitation, there is a clear trade-off between preserving temporal information (more frames) and spatial details (higher resolution).
For example, to manage a reasonable input length, prior work~\cite{damonlpsg2023videollama,lin2024videollavalearningunitedvisual,luo2023valley,videollmonlinechen2024} compressed each frame into a single video token. While this might be sufficient for downstream tasks that only require global content like action classification~\cite{kay2017kineticshumanactionvideo,caba2015activitynet}, it is inadequate for tasks that require fine-grained details, such as video dense captioning~\cite{Youcook2ZhXuCoAAAI18}, highlights detection~\cite{QVHighlights2021}, temporal action localization~\cite{kay2017kineticshumanactionvideo}, and video temporal grounding~\cite{gao2017talltemporalactivitylocalization}. These types of tasks are time-sensitive video understanding (TSV) tasks, as they require 
detailed object-level information, including classes, attributes, and spatial relationships, to be preserved across frames.
% \footnote{Please refer to the project page at \url{https://intelailabpage.github.io/2025/12/08/go-video.html}.}

\begin{figure}[t!]
    \centering
    \includegraphics[width=\linewidth]{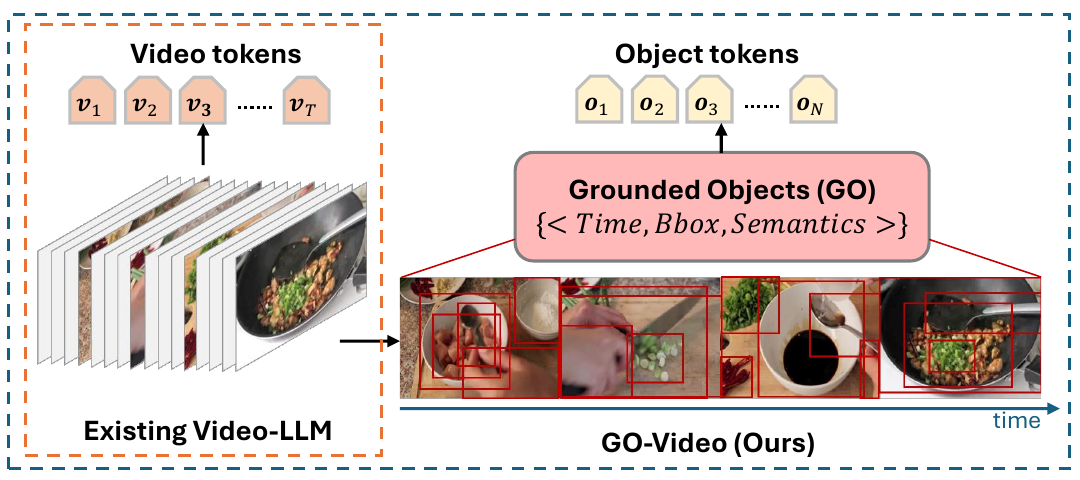}
    \vspace{-12pt}
    \caption{Video tokens in existing Video-LLMs~\cite{litahuang2024} are highly compressed and compromise on spatial information. We harness GO information in sparsely sampled video frames as a supplement to improve the time-sensitive video understanding (TSV) tasks.
    }
    \label{fig:teaser}
    \vspace{-8pt}
\end{figure}

In this work, we investigate the task of TSV and hypothesize that the performance of TSV can be improved when the object-level information is preserved.
Since recent Video-LLMs~\cite{lin2024videollavalearningunitedvisual, damonlpsg2023videollama,cheng2024videollama2advancingspatialtemporal, maaz2024videochatgptdetailedvideounderstanding,TimeChatRen2023} do not harness such object-level information either during training or inference, we first conduct a preliminary study on this problem with an existing Video-LLM, namely LITA~\cite{litahuang2024}. More specifically, we convert the object annotations in the ActivityNet-Captions~\cite{activityNetCaptionskrishna2017dense} dataset into a textual description encompassing the object classes, locations, and the timestamps where the objects appear.
These object descriptions across frames are then passed to the Video-LLM to perform downstream tasks. Initial results using this approach demonstrate an improvement in the model's performance, suggesting the need to integrate object information into existing Video-LLMs. 
However, this gain comes with the cost of having an excessive amount of extra tokens. For example, when introducing the object description that covers the object class, bounding box, and timestamp of the objects, the model consumes around 1,400 additional tokens during inference, compared to vanilla LITA. Furthermore, the preliminary study also shows that the gain diminishes significantly when the text description contains inaccurate object classes or misaligned bounding boxes. This finding motivated us to investigate an effective and efficient method for utilizing object-level information in temporal understanding tasks. 

To address this, we propose {\bf GO-Tokenizer}, a lightweight module for existing Video-LLMs that encodes the object semantics, spatial positions, and temporal cues of the {\bf{grounded objects (GO)}} in video frames into compact object tokens, as illustrated in Figure~\ref{fig:teaser}.
For each object in a frame, we introduce ROI patch pooling to average the patch features within the given object bounding box, where the patch features combines a semantic-rich visual feature map from a pretrained encoder~\cite{YOLOWorldCheng2024}) and location-aware positional embeddings. This pooling mechanism inherently captures the semantic and spatial information. The resulting feature is projected into a latent space shared with text tokens and enriched with a time-aware positional encoding to capture when the object appears.
Each object is thus efficiently represented as a single object token, which is passed to the LLM alongside text and video tokens.
The GO-Tokenizer module is seamlessly integrated with existing Video-LLMs, referred to as GO-Video models, which is optimized end-to-end with low-rank adaptation~\cite{hu2022lora} applied in the LLM component.
During inference, where GO annotations are unavailable, we uniformly sample sparse frames from the video, leveraging the redundancy in adjacent frames.
We employ an off-the-shelf object detector~\cite{YOLOWorldCheng2024,deticzhou2022detecting} to extract GO information in these frames, and encode them with GO-Tokenizer into object tokens. This strategy readily ensures precise temporal grounding.

To demonstrate the efficacy of the GO-Tokenizer, we integrate it into two leading Video-LLMs, TimeChat~\cite{TimeChatRen2023} and LITA~\cite{litahuang2024}, resulting models referred to as GO-TimeChat and GO-LITA respectively, where the former is evaluated on the dense video captioning task using YouCook-2~\cite{Youcook2ZhXuCoAAAI18} dataset and the latter is evaluated on the task of reasoning temporal localization using the ActivityNet-RTL~\cite{litahuang2024} dataset. In both tasks, models equipped with the GO-Tokenizer outperform (1) their vanilla version and (2) their vanilla version trained and inferred with object text descriptions. Additionally, we conduct ablation studies considering different configurations during inference, including the use of different object detectors and the number of objects per video frame. Under all configurations, the proposed GO-Tokenizer outperforms the baselines using a more robust and compact representation to encode object information.

Overall, this work investigates how to effectively and efficiently utilize GO information for time-sensitive video understanding, making the following three contributions. 
First, we demonstrate that, without modifying the underlying architecture, existing Video-LLMs can be improved by incorporating object information at varying levels of granularity, such as class labels and spatial grounding, for training and testing at the text prompt level.
Second, we propose to utilize the object-level visual features from existing object detectors and train learnable grounded object (GO) tokens to efficiently encode object information.
Finally, we show that the proposed GO-Tokenizer consistently outperforms the baselines on the TSV tasks when applied to two different Video-LLMs, LITA and TimeChat.

%% file: 2_related_works.tex
\section{Related Work}

\paragraph{Large Language Models for Video Understanding.}
The advent of large language models (LLM) pre-trained on extensive datasets notably increases the capability of handling various tasks using prompts without requiring fine-tuning. Several studies have been made about the utilization of LLMs in the computer vision field, such as LLaVA \cite{liu2023llava}, GPT-4V\cite{yang2023dawnlmmspreliminaryexplorations}, and Visual-ChatGPT~\cite{Wu2023VisualCT}.
LLMs integrated with video understanding opened up completely new ways of interacting with videos. These models are equipped with the power of complex multimodal reasoning, interacting with textual prompts in ways more natural for humans. 
In recent years, there has been a surge of models for such Video-LLM tasks. The survey paper~\cite{tang2024videounderstandingLL} is an excellent reference to the Video-LLM field as a whole. Typically, most of these models utilize open-source LLMs like LLaMA\cite{touvron2023llamaopenefficientfoundation} or Vicuna\cite{chiang2023vicuna} as the backbone. Recent Video-LLM models significantly outperform previous methods on several video understanding tasks, ranging from video classification to video question-answering, with the aim of near-human level performance for video interpretation capability. 

% \vspace{-5pt}
\paragraph{Time-Sensitive Video Understanding Tasks.}
Video understanding encompasses a range of tasks, each with distinct applications and benchmarks. These tasks can be broadly grouped into three categories.
The first involves holistic understanding, including video classification~\cite{abu2016youtube}, action recognition~\cite{kuehne2011hmdb}, multi-modal retrieval~\cite{gabeur2020multi}, video captioning~\cite{xu2016msr}, and video-to-textual summarization~\cite{Hua2024V2XumLLMCV}, etc.
The second category, containing time-sensitive video understanding (TSV) tasks, focuses on temporal understanding and includes tasks such as video highlights~\cite{QVHighlights2021}, condensing long video to short duration~\cite{qiu2021condensing}, temporal action or event localization~\cite{Qian2024MomentorAV}, moment retrieval~\cite{10350951}, dense video captioning~\cite{Zhou2024StreamingDV}.
Finally, the third category centers on spatial-temporal understanding and is associated with object-level tracking~\cite{Wang2024ElysiumEO}, re-identification~\cite{Yang2024MLLMReIDML}, segmenting video objects~\cite{yan2024visa}, etc. 
This work concentrates on the second category, which is TSV.
While objects provide valuable cues for interpreting scene semantics, we observe that state-of-the-art Video-LLM models~\cite{TimeChatRen2023,huang2024lita, damonlpsg2023videollama,li2024videochatchatcentricvideounderstanding,luo2023valley, cheng2024videollama2advancingspatialtemporal, maaz2024videochatgptdetailedvideounderstanding} developed for TSV do not explicitly leverage such information.
To assess the utility of grounded object information for this task, we augment existing Video-LLMs with textual descriptions of detected objects, and further propose a token-efficient representation strategy for these objects, leading to improved performance on TSV tasks.

% \vspace{-5pt}
\paragraph{Large Multimodal Models with Object Information.}
The core idea in Multimodal Large Language Models (MLLMs) lies in vision tokenization, which transforms visual signals into feature representations apt for pre-trained LLMs.  
Recent approaches for MLLMs introduced several ways to leverage explicit entity-level information for image-based reasoning tasks. 
RegionGPT~\cite{guo2024RGPT} enables users to specify a region of interest, and such placeholders are subsequently replaced with semantic region-level embeddings that are later consumed by the language model.
VCoder~\cite{jain2024vcoder} explores off-the-shelf object perception modalities such as segmentation or depth maps as additional control signals and projects this information into the LLM input space. 
Another recent work, MG-LLaVA~\cite{mgllava}, introduced concept-level image-text alignment via semantically equivalent visual tokens.
While these methods focus on visual tokenization strategies for image-level reasoning, they design tokenization schemes tailored to their specific tasks with static images rather than videos, where temporal dynamics pose critical challenges.
Although GO-Tokenizer is also designed to be integrated into a pretrained LLM, it is tailored for video understanding, especially for TSV tasks.

%% file: 3_method.tex
\section{Preliminaries}
\paragraph{Video-LLM architecture.}
LLMs have been widely adapted to different modalities due to their established efficacy in sequence modeling and remarkable performance in predicting text tokens. 
In the visual modality, a popular approach is to represent an image with a series of visual patch tokens~\cite{dosovitskiy2020vit}, and these visual tokens are projected to a latent space that is shared with the text tokens~\cite{liu2023llava}. The LLM then ingests the combined text tokens and visual tokens and generates text tokens for various downstream tasks.
Similarly, video large language models (Video-LLMs) extract video tokens with a video encoder (e.g., Video-Qformer~\cite{TimeChatRen2023} or ViT~\cite{dosovitskiy2020vit}) from the given video ${\bf V}\in\mathbbm{R}^{T\times H\times W\times 3}$, where $T$ denotes the number of frames, and $H, W$ are the spatial dimensions of the video frame.
In summary, Video-LLMs consume two types of input tokens, video tokens ${\bf v}_{1...K}$ and text tokens ${\bf t}_{1...L}$, with $K$ and $L$ denoting the token length of the respective modality.

\paragraph{Advancing Video-LLMs for TSV.}
Time-sensitive video understanding (TSV) capability of Video-LLMs is typically empowered by instruction tuning~\cite{zhang2024llavanext-video,liu2023llava} the model on large, annotated video datasets with temporal labels and event-related queries. This enables the model to understand and predict the temporal relationships between events, such as when they start, end, or overlap. For instance, in event localization, the instruction could be {\it ``When does [event] happen in the video? Answer the question only using start and end timestamps,"} with the model providing the corresponding timestamps. Similarly, for dense video captioning, the instruction might be {\it ``Detect and report the start and end timestamps of activity events in the video, along with descriptions."}, with the model generating a series of narrations with their timestamps.
Since the output is natural language, the model is usually trained using next-token prediction as its optimization objective.

\begin{figure*}[t!]
    \centering
    \begin{minipage}[h]{0.38\linewidth}
        \centering
        \resizebox{\linewidth}{!}{
        \setlength{\tabcolsep}{8pt}
        \begin{tabular}{l|cc}
        \toprule
            GO at Inference & mIOU & P@0.5  \\ \midrule
            n/a & 23.04 & 18.78 \\ \hline
            Text {\it (Class)} & 23.65 & 18.08 \\
            Text {\it (Class+Time)} & 26.47 & 20.47 \\
            Text {\it (Class+Time+Bbox)} & 27.63 & 23.71 \\
        \bottomrule
        \end{tabular}
        }
        \vspace{-5pt}
        \captionof{table}{Using different levels of ground truth GO information as context to evaluate LITA-13B~\cite{litahuang2024} on ActivityNet-RTL-GO.}
        \label{tab:lita_go-text}
    \end{minipage}
    \hfill
    \begin{minipage}[h]{0.27\linewidth}
        \centering
        \resizebox{\linewidth}{!}{
        \setlength{\tabcolsep}{4pt}
        \begin{tabular}{l|cc}
        \toprule
            Flip Ratio & mIOU & P@0.5  \\ \midrule
            10\% & 27.07	& 23.25 \\
            20\% & 25.89	& 22.63 \\
            50\% & 24.63	& 20.13 \\
        \bottomrule
        \end{tabular}
        }
        \vspace{-5pt}
        \captionof{table}{Flipping a percentage of object class labels in each video into randomly sampled classes.}
        \label{tab:flip_label}
    \end{minipage}
    \hfill
    \begin{minipage}[h]{0.31\linewidth}
        \centering
        \resizebox{\linewidth}{!}{
        \setlength{\tabcolsep}{4pt}
        \begin{tabular}{l|cc}
        \toprule
            Shift & mIOU & P@0.5  \\ \midrule
            $0.01\times(H, W)$ & 27.37 & 21.13 \\
            $0.02\times(H, W)$ & 26.42 & 19.01 \\
            $0.05\times(H, W)$ & 24.71 & 17.14 \\
        \bottomrule
        \end{tabular}
        }
        \vspace{-5pt}
        \captionof{table}{Shifting bounding box locations by varying amounts with respect to the image height ($H$) and width ($W$) in pixels.}
        \label{tab:bbox_shift}
    \end{minipage}
    \vspace{-9pt}
\end{figure*}

\section{Motivation: Can object grounding in Video-LLMs help TSV?}\label{sec:prelim}

Due to computational limitations and the complexity of modeling long sequences, the existing LLMs can only process a finite number of tokens, and such a constraint leads to the trade-off between spatial and temporal resolutions~\cite{xu2024slowfast,wu2024freeva} for Video-LLMs. For example, to model the long-sequence video (i.e., large $T$), the size of each video frame is reduced (i.e., small $H, W$), which leads to fewer video tokens per frame. Prior works~\cite{damonlpsg2023videollama, li2024videochatchatcentricvideounderstanding,luo2023valleyvideoassistantlarge} even represent each frame with only one video token (i.e., $K=T$).
Since the compression in spatial dimension indicates the loss of fine-grained spatial details, this hinders the Video-LLMs from excelling in time-sensitive downstream tasks that require fine-grained information in each frame, such as reasoning temporal localization~\cite{litahuang2024} and dense video captioning~\cite{Youcook2ZhXuCoAAAI18}.
Recently, LITA~\cite{litahuang2024} introduced the {\it Slow-Fast} tokens for this trade-off, with {\it Slow} and {\it Fast} tokens that favor the spatial and time resolution, respectively.  

In this paper, we explore an orthogonal direction, augmenting the highly condensed video tokens on a more abstract level. We hypothesize that leveraging the sparse but informative {\bf grounded objects (GO)} in video frames could improve the performance of TSV tasks.
We aim to answer two questions: (1) Can providing GO information in Video-LLMs enhance TSV? (2) What levels of GO information are needed?
Object grounding in the wild, however, may contain some noise in the class label and bounding box location space. To disentangle this factor from our hypothesis, we construct a subset of ActivityNet-RTL~\cite{litahuang2024} evaluation set where object entity annotations (e.g., object class, bounding box location, etc.) are available from~\cite{ActivityNetEntitieszhou2019grounded}, namely {\bf ActivityNet-RTL-GO}. The dataset sparsely labels an average of $10.43$ objects per video and $2.41$ objects per frame, with object labels present in $4.45$ frames on average.
In this section, we conduct empirical studies on these questions using LITA~\cite{litahuang2024}, a state-of-the-art Video-LLM for TSV, and evaluate its temporal localization performance on the ActivityNet-RTL-GO dataset, with different levels of GO information provided at inference time.

% \vspace{-5pt}
\subsection{Augmenting Video-LLMs with GO information}
Object grounding in videos involves several dimensions, including the object category {\it (Class)}, the timestamp where the objects are visible {\it (Time)}, and the spatial locations in the frames {\it (Bbox)}. To incorporate GO into Video-LLMs, we encode this information into text descriptions as additional context prepended to the text instructions during inference. 
We provide GO information in various granularities to Video-LLMs as distinct variants to examine which levels are essential for temporal understanding, i.e.,
\begin{itemize}
    \setlength{\itemsep}{1.5pt}
    \item {\bf Class:} {\it ``\bracket{Obj} Objects in this video are: man, window, ...\bracket{/Obj}"}.
    \item {\bf Class+Time:} {\it ``\bracket{Obj} Each object is provided with its timestamp and class label in the format of \bracket{time, class label}. Here are the objects: \bracket{91.2 second, man}, ...\bracket{/Obj}"}.
    \item {\it {\bf Class+Time+Bbox:} ``\bracket{Obj} Each object bounding box is provided with its timestamp and class label in the format of \bracket{time, (x1,y1,x2,y2), class label}. Here are the objects: \bracket{91.2 second, (0.0001, 0.1715, 0.0806, 0.3784), man}, ...\bracket{/Obj}"}.
\end{itemize}

Table~\ref{tab:lita_go-text} presents the temporal localization performance of LITA with these variants, with the first row showing the baseline performance without any GO information provided. The mIOU calculates the average intersection-over-union (IOU) between the predicted and ground truth time segments, whereas P@0.5 measures precision at a 0.5 IOU threshold.
When introducing GO information, the results demonstrate a noticeable improvement over the baseline, with greater gains observed as more fine-grained GO information is provided.
This supports the hypothesis that augmenting video tokens with GO can improve the TSV capability of Video-LLMs. However, assuming perfect object grounding at test time is unrealistic.
\begin{figure*}[t!]
    \centering
    \includegraphics[width=0.97\linewidth]{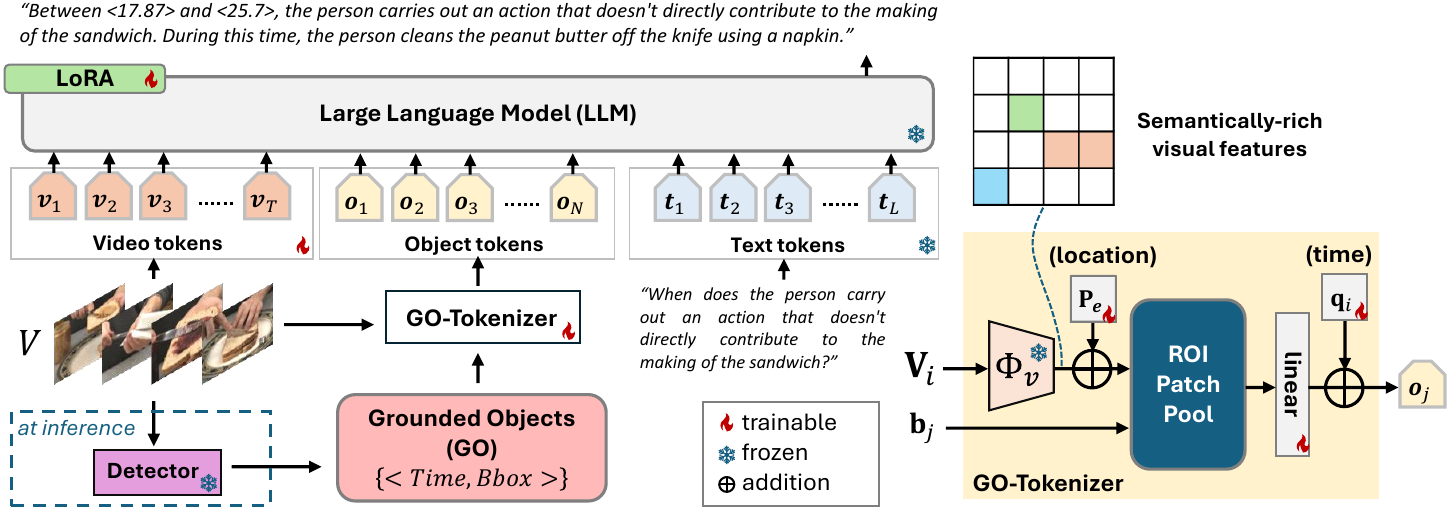}
    \vspace{-5pt}
    \caption{(Left) Model architecture of GO-Video. The LLM input space is augmented with the object tokens extracted by the GO-Tokenizer. GO information is extracted with off-the-shelf object detectors during inference. (Right) GO-Tokenizer extracts the object semantics, locations and time information of each object into a single object token.}
    \label{fig:model}
    \vspace{-7pt}
\end{figure*}

\subsection{Analysis of Object Grounding Perturbations}
To evaluate the robustness of incorporating GO information as text instructions, we simulate the perturbations in object grounding by introducing noise to ground truth class labels and bounding box locations. Perturbed GO information is then applied to the {\it Class+Time+Bbox} approach from the previous section.
Note that time annotation is always precise during inference, as object grounding is obtained by running object detection on the sampled video frames. We elaborate the perturbations considered below.

% \vspace{-11pt}
\paragraph{Class flipping.}
This operation randomly selects $x\%$ of the objects per video, and replaces their class labels with randomly chosen labels from the dataset, simulating classification errors in object detection. We report the average performance for three random seeds.
% \vspace{-11pt}
\paragraph{Bounding box shift.}
The object localization error is simulated by shifting ground truth bounding box positions by a percentage of the image's height and width, allowing control over the noise in object detection.

Table~\ref{tab:flip_label} and Table~\ref{tab:bbox_shift} show that providing GO information simply as text is not robust to the perturbations in object grounding, as performance degrades at higher noise levels.
Another side effect of this approach is the inefficiency in the token length for representing an object. 
For instance, {\it Class+Time+Bbox} requires roughly 42 additional tokens per object, with the total overhead increasing proportionally to the number of objects.
This raises the question of what is the most effective and generalizable manner to represent the GO information, so that it can generalize across visual domains while keeping the input sequence to Video-LLMs within a manageable length.
Inspired by the preliminary experiments, we introduce the notion of object tokens ${\bf o}_{1...N}$, representing the GO information for $N$ objects at a more abstract level. This is a new type of input token in Video-LLMs to supplement the video and text tokens. The details of the GO tokenization are presented in the next section.

\section{Grounded Object Tokenization}
To encode GO information with a more efficient and generalizable representation to Video-LLMs, we propose {\bf GO-Tokenizer} as a lightweight add-on module to abstract the GO information into object tokens. We refer to the Video-LLM architecture supplemented with GO-Tokenizer as {\bf GO-Video}, as illustrated in the left of Figure~\ref{fig:model}.

\subsection{Architecture}\label{sec:go-tokenizer}
Given an image ${\bf V}_i\in\mathbbm{R}^{H\times W\times 3}$ sampled at $i$-th video frame and a bounding box coordinate ${\bf b}_j\in\mathbbm{R}^4$ associated with an object within the image, GO-Tokenizer extracts an object token ${\bf o}_j\in\mathbbm{R}^{d_t}$, which contains the object semantics, bounding box locations and time information in a single token. 
More specifically, given an image ${\bf V}_i\in\mathbbm{R}^{H\times W\times 3}$, we leverage the YOLO-World backbone~\cite{YOLOWorldCheng2024} as a pre-trained visual encoder $\Phi_v$, to extract the frame-level feature map.
YOLO-World is selected because it is a lightweight CNN-based architecture and it is trained with region-based image-text alignment using the text encoder from CLIP~\cite{ClipRadford2021LearningTV}. Note that other feature extractors like the ViT in vanilla CLIP are also feasible, while it does not have image-text alignment at the region level but the image level~\cite{ClipRadford2021LearningTV}, which tends to lead to inferior performance (as shown in the appendix).
To encode the location information, we add a patch positional embedding ${\bf P}_e$ to the frame-level feature map extracted by $\Phi_v$, i.e.
\begin{align}
    {\bf F}_i = \Phi_v({\bf V}_i) + {\bf P}_e
\end{align}
where $\Phi_v({\bf V}_i),{\bf P}_e\in\mathbbm{R}^{N_p\times N_p\times d_v}$, $N_p$ denote the number of patches, and $d_v$ the dimension of the visual feature. Note that we keep $\Phi_v$ frozen in both training and inference.  

\begin{figure}[t!]
    \centering
    \includegraphics[width=0.8\linewidth]{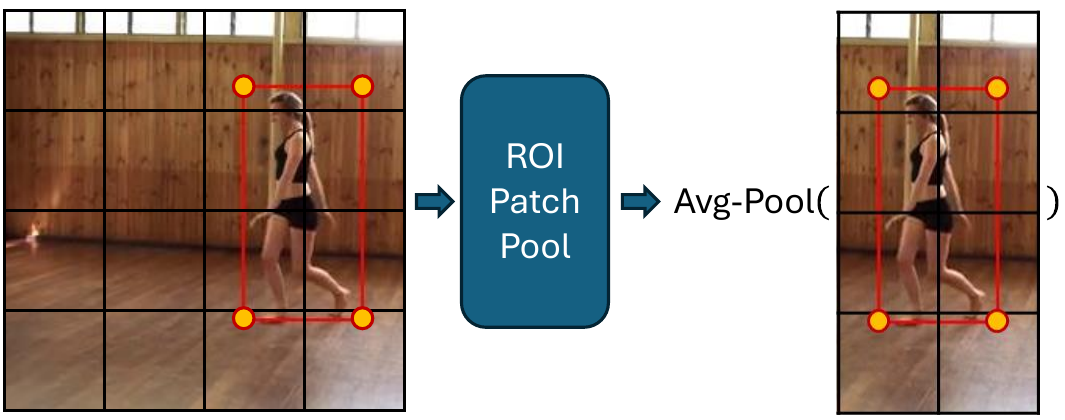}
    \caption{{\bf ROI-Patch-Pool.} All the patches covered by the bounding box are average-pooled.}
    \label{fig:patch_pool}
    \vspace{-7pt}
\end{figure}

To acquire semantically-rich object features, we use the given object bounding box ${\bf b}_j$ to crop the ROI (region of interest) in the frame-level feature map ${\bf F}_i$. This is implemented with a $\textit{ROI-Patch-Pool}(\cdot)$ that extracts the object ROI feature ${\bf h}_j$ by averaging over all the patches within the patches covered by the bounding box, as illustrated in Figure~\ref{fig:patch_pool}. The $\textit{ROI-Patch-Pool}$ operation can be formulated as
\begin{align}
    {\bf h}_j = \textit{ROI-Patch-Pool}({\bf F}_i, {\bf b}_j).
\end{align}
Since we are performing a pooling operation in the feature map, the architecture becomes more robust to minor errors in bounding box predictions. 
Finally, to project the ROI feature to the same latent space of the text tokens, a linear projection layer ${\bf W}_o\in\mathbbm{R}^{d_v\times d_t}$ is trained and a time-aware positional embedding ${\bf q}_i$ is added, i.e.,
\begin{align}
    {\bf o}_j = {\bf W}_o^T {\bf h}_j + {\bf q}_i,
\end{align}
where ${\bf q}_i\in\mathbbm{R}^{d_t}$ represents the $i$-th frame and ${\bf o}_j$ is the object token that represents an object in the $i$-th frame. 
The overall architecture of GO-Tokenizer is summarized in the right of Figure~\ref{fig:model}.

\subsection{Training and Inference}
To demonstrate the generality of the GO-Tokenizer, 
we integrate it to two state-of-the-art Video-LLMs tuned for TSV tasks, LITA~\cite{litahuang2024} and TimeChat~\cite{TimeChatRen2023}, resulting in GO-LITA and GO-TimeChat, respectively. While they employ different model architectures to extract video tokens (e.g., ViT and Q-Former), GO-Tokenizer remains a versatile module applicable to different Video-LLM models.

For each GO-Video model (i.e., GO-LITA/GO-TimeChat), the model extracts video tokens and text tokens following the original setting, while extracting object tokens with GO-Tokenizer.
The LLM consumes the concatenation of three types of input tokens
\begin{align}
    {\bf v}_{1...K} \oplus {\bf o}_{1...N} \oplus {\bf t}_{1...L}
\end{align}
where $\oplus$ denotes sequence concatenation.
We optimize the Video-LLM with GO-Tokenizer (with trainable parameters: ${\bf W}_o$, ${\bf q}$, ${\bf P}_e$) in an end-to-end manner with the language modeling loss in their original settings. Low-rank adaptation~\cite{hu2022lora} is applied to the LLM component, keeping the majority of its parameters fixed. Detailed settings for each model are elaborated in the experiment section.

During inference, we extract GO information on-the-fly by applying an off-the-shelf object detector~\cite{YOLOWorldCheng2024,deticzhou2022detecting} on sparse frames uniformly sampled from the video. To reduce noisy predictions, we set a confidence threshold $\delta$ and
only keep the top-$k$ object predictions above that threshold. Thus, for a video with $F$ sampled frames, the number of object tokens $N$ is bounded by $F\times k$. In this paper, we set $F=8$ and $k=5$ if not specifically noted. The extracted GO information is then encoded by the GO-Tokenizer into object tokens.

%% file: 4_experiments.tex
\section{Experiments}
\newcommand{\xmark}{\ding{55}}
\subsection{Experimental Setup}\label{sec:setting}
\paragraph{GO-LITA.}
We obtain object annotations from ActivityNet-Entities~\cite{ActivityNetEntitieszhou2019grounded} for both ActivityNet-RTL~\cite{litahuang2024} and ActivityNet-Captions~\cite{activityNetCaptionskrishna2017dense}. We finetune the LITA-13B model jointly with the former dataset on reasoning temporal localization (RTL) and the latter on dense video captioning and event localization tasks.
The ActivityNet-RTL~\cite{litahuang2024} dataset comprises 10,009/160 training/evaluation videos with 33,557/229 RTL question-answer pairs generated by GPT-4. The RTL task demands advanced reasoning of the video content to determine the event's timing and offer an explanation.
We evaluate the model on the full ActivityNet-RTL~\cite{litahuang2024} dataset and provide the results on ActivityNet-RTL-GO in the appendix.
We report the mIOU and P@0.5 metrics for temporal localization. For the explanation evaluation, since we don't have access to the GPT-4 API, we evaluate the relative score (Rel. Score) metric~\cite{litahuang2024} with Llama-3-8B-Instruct~\cite{llama3modelcard}.
\vspace{-5pt}
\paragraph{GO-TimeChat.}
Similar to the GO-LITA setting, we enrich ActivityNet-Captions~\cite{activityNetCaptionskrishna2017dense} with bounding box annotations~\cite{ActivityNetEntitieszhou2019grounded} and finetune the vanilla TimeChat model with the dense video captioning task.
We evaluate our approach on the YouCook-2 dataset~\cite{Youcook2ZhXuCoAAAI18}, which comprises approximately 1,790 raw cooking video clips. The model is tasked with identifying around eight essential cooking steps and generating descriptions that accurately reflect the visual content. We use only the validation portion of the dataset for the zero-shot evaluation and 
report performance using CIDEr, F1 score, METEOR, and SODAc metrics.

% \vspace{-8pt}
\paragraph{Implementation.}
We build the codebase atop LITA~\cite{litahuang2024} and TimeChat~\cite{TimeChatRen2023} for GO-LITA and GO-TimeChat, respectively.
All experiments were trained with 4 NVIDA-A100 GPUs. More details are reported in the appendix.

\begin{figure}[t!]
    \centering
    \resizebox{\linewidth}{!}{
    \setlength{\tabcolsep}{2.5pt}
    \begin{tabular}{l|c|ccccc}
    \toprule
         Model & GO Training & mIOU & P@0.5 & Rel. Score \\ \midrule
         LITA & \xmark & 28.28 & 26.40 &  54.7 \\ 
         LITA + Text ({\it Class+Time+Bbox}) & \xmark & 28.07 & 25.83 & 54.3 \\ \midrule
         \rowcolor{lightgray} \multicolumn{5}{c}{\it Finetuned on ActivityNet-Captions \& ActivityNet-RTL} \\
         LITA  & \xmark & 28.72 & 24.53 & 54.7 \\
         LITA + Text ({\it Class+Time+Bbox}) & \xmark & 28.82 & 25.10 & 55.6 \\ \hline
         GO-LITA & \checkmark & {\bf 31.52} & {\bf 28.07} & {\bf 59.0} \\
         \bottomrule
    \end{tabular}
    }
    \vspace{-5pt}
    \captionof{table}{Evaluating LITA-13B on the ActivityNet-RTL dataset with/without GO-Tokenizer. YOLO-World~\cite{YOLOWorldCheng2024} is adopted at inference time to extract the GO information.}
    \label{tab:GO-LITA}
    \vspace{8pt}
    \centering
    \resizebox{\linewidth}{!}{
    \setlength{\tabcolsep}{3pt}
    \begin{tabular}{l|c|ccc}
        \toprule
         Model & GO Training & CIDEr & SODAc & F1 \\ \midrule
         Valley~\cite{luo2023valley} & \xmark & 0.0 & 0.1 & 1.5 \\
         Video-LLaMA~\cite{damonlpsg2023videollama} & \xmark  & 0.0 & 0.0 & 0.1 \\
         TimeChat~\cite{TimeChatRen2023} & \xmark  & 3.4 & 1.2 &  12.6 \\ \midrule
         \rowcolor{lightgray} \multicolumn{5}{c}{\it Finetuned on ActivityNet-Captions} \\
         TimeChat & \xmark  & 3.4 & 1.1 & 11.3 \\
         TimeChat + Text ({\it Class+Time+Bbox}) & \xmark  & 2.4 & 0.8 & 10.6 \\ \hline
         GO-TimeChat & \checkmark & \textbf{3.9} & \textbf{1.4}  & \textbf{18.5} \\
        \bottomrule
    \end{tabular}
    }
    \vspace{-5pt}
    \captionof{table}{Zero-shot evaluation on YouCook-2~\cite{Youcook2ZhXuCoAAAI18} dataset for dense video captioning. GO information is extracted with Detic~\cite{deticzhou2022detecting} at inference. All the LLMs have approximately 7B parameter size.}
    \label{tab:GO-TimeChat}
    \vspace{-12pt}
\end{figure}

% \vspace{-5pt}
\subsection{Comparisons to Vanilla Video-LLMs}
% \vspace{-2pt}
Table~\ref{tab:GO-LITA} and Table~\ref{tab:GO-TimeChat} present the main results of utilizing GO-Tokenizer to encode object-level information, compared to the vanilla Video-LLMs and their text-based baselines.
We evaluate two state-of-the-art Video-LLM models for the TSV task, LITA~\cite{litahuang2024} and TimeChat~\cite{TimeChatRen2023}, whose original models were trained on broader datasets (e.g., TimeIT~\cite{TimeChatRen2023}) alongside ActivityNet.
For fair comparison, we fine-tune all models exclusively using ActivityNet, as it provides the necessary object entity annotations. The upper blocks of the tables report the performance of the published checkpoints for reference.
Note that only GO-LITA and GO-TimeChat leverage the bounding box annotations during training.
Compared to vanilla LITA or TimeChat, introducing GO-Tokenizer to these models significantly boosts the performance on all the metrics across datasets and models.
Interestingly, vanilla TimeChat shows no improvement when GO information is provided as text ({\it Class+Time+Bbox}) during inference. This may be attributed to its zero-shot evaluation setting or limitations in context length, as encoding object-level details in text is token-inefficient.
LITA, with twice the context length of TimeChat, is less affected by this constraint.
In contrast, GO-Tokenizer offers a more scalable and generalizable approach to integrating object-level information to Video-LLMs.

\begin{figure}[t!]
    \centering
    \setlength{\tabcolsep}{2pt}
    \resizebox{\linewidth}{!}{
    \begin{tabular}{{l|cc|ccc}}
        \toprule
         Model & Go Training & Tokens/GO & CIDEr & SODAc & F1  \\ \midrule
        TimeChat & \xmark & 0 & 3.4 & 1.1 & 11.3 \\ \midrule
        TimeChat + Text ({\it Class}) & \checkmark & 4 & 1.8	& 0.8 & 16.2  \\ 
        TimeChat + Text ({\it Class+Time}) & \checkmark & 12 & 1.8	& 0.9 & 16.8  \\
        TimeChat + Text ({\it Class+Time+Bbox}) & \checkmark & 42 & 1.6  & 0.8 & 13.0 \\ \midrule
        GO-TimeChat & \checkmark & 1 & {\bf 3.9} & {\bf 1.4} & {\bf 18.5} \\
        \bottomrule
    \end{tabular}
    }
    \vspace{-4pt}
    \captionof{table}{Comparisons to baselines that represent GO as text descriptions with TimeChat. Detic~\cite{deticzhou2022detecting} is used to extract GO information at inference.}
    \label{tab:timechat-baseline}
    \vspace{5pt}
    \resizebox{\linewidth}{!}{
    \begin{tabular}{l|c|ccc}
        \toprule
         Model & Go Training & mIOU & P@0.5 & Rel. Score \\ \midrule
         LITA & \xmark & 28.72 & 24.53 & 54.7 \\
         LITA + Text ({\it Class+Time+Bbox}) & \checkmark & 30.05	& 27.97 & 55.5 \\
         GO-LITA & \checkmark & {\bf 31.52} & {\bf 28.07} & {\bf 59.0} \\
        \bottomrule
    \end{tabular}
    }
    \vspace{-4pt}
    \captionof{table}{Comparisons to baselines that represent GO as text descriptions with LITA. YOLO-World~\cite{YOLOWorldCheng2024} is used to extract GO information at inference.}
    \label{tab:lita-baseline}
    \vspace{5pt}
    \centering
    \resizebox{\linewidth}{!}{
    \begin{tabular}{l|cc|cc}
    \toprule
        Model & Object detector & Go Training & CIDEr & F1 \\ \midrule
        TimeChat & n/a & \xmark & 3.4 & 11.3 \\ \midrule
        TimeChat + Text ({\it Class+Time+Bbox})  & Detic & \checkmark & 1.6 & 13.0 \\
        GO-TimeChat & Detic  & \checkmark & 3.9 & 18.5\\ \midrule
        TimeChat + Text ({\it Class+Time+Bbox}) & YOLO-World  & \checkmark &  1.7 & 14.3 \\
        GO-TimeChat & YOLO-World  & \checkmark & 4.4 & 18.5\\
        \bottomrule
    \end{tabular}
    }
    \vspace{-5pt}
    \captionof{table}{Ablations on object detectors. Zero-shot results on YouCook-2. The performance of the GO-Tokenizer generalizes across the selection of object detectors.}
    \label{tab:objec_detector_ablade}
    \vspace{-5pt}
\end{figure}

\subsection{Comparisons to Training with GO as Text}\label{sec:baseline}
Since the results of both GO-LITA and GO-TimeChat have shown improvements on TSV tasks when finetuning Video-LLMs with ground truth object annotations, it is also feasible to incorporate GO information as text descriptions in both finetuning and testing phases, using the formulation described in section~\ref{sec:prelim}. We implement these variants as baselines, denoted as [Video-LLM] + Text 
({\it Class}/{\it Class+Time}/{\it Class+Time+Bbox}).
Note that all these baselines follow the same training/evaluation setting as GO-LITA and GO-TimeChat, respectively.  

Table~\ref{tab:timechat-baseline} compares the zero-shot performance when the models are trained on ActivityNet-Captions and evaluated on YouCook-2. While the temporal localization performance (F1) improves for some baselines, the caption quality (CIDEr, SODAc) decreases drastically, possibly because these models are not resilient to object detection noise, as shown in section~\ref{sec:prelim}, or the excessive tokens introduced by the GO text description exceeds the context length limitation.
Note that the number of additional tokens allocated per GO in these baselines is much higher than in GO-Tokenizer, which increases proportionally with the object attributes considered (i.e. the token length of {\it Class+Time} is greater than that of {\it Class}). Given that the model's maximum sequence length remains fixed, this may prohibit users from issuing a more complex query.
For example, sampling 16 frames per video and assuming only one object per frame would consume $42 \times 16$ tokens.
In contrast, GO-Tokenizer abstracts the object information into a single token, offering the scalability as there are more objects in each frame.

Table~\ref{tab:lita-baseline} presents the comparison of GO-LITA to the most informative baseline ({\it Class+Time+Bbox}). Unlike TimeChat, the temporal localization metrics (mIOU and P@0.5) show improvement with both the text-based baseline and the GO-Tokenizer. GO-LITA further enhances the improvement over the baseline on the Rel. score metric.

\vspace{2pt}
\subsection{Ablation Studies}

\begin{figure*}[t!]
    \centering
    \includegraphics[width=\linewidth]{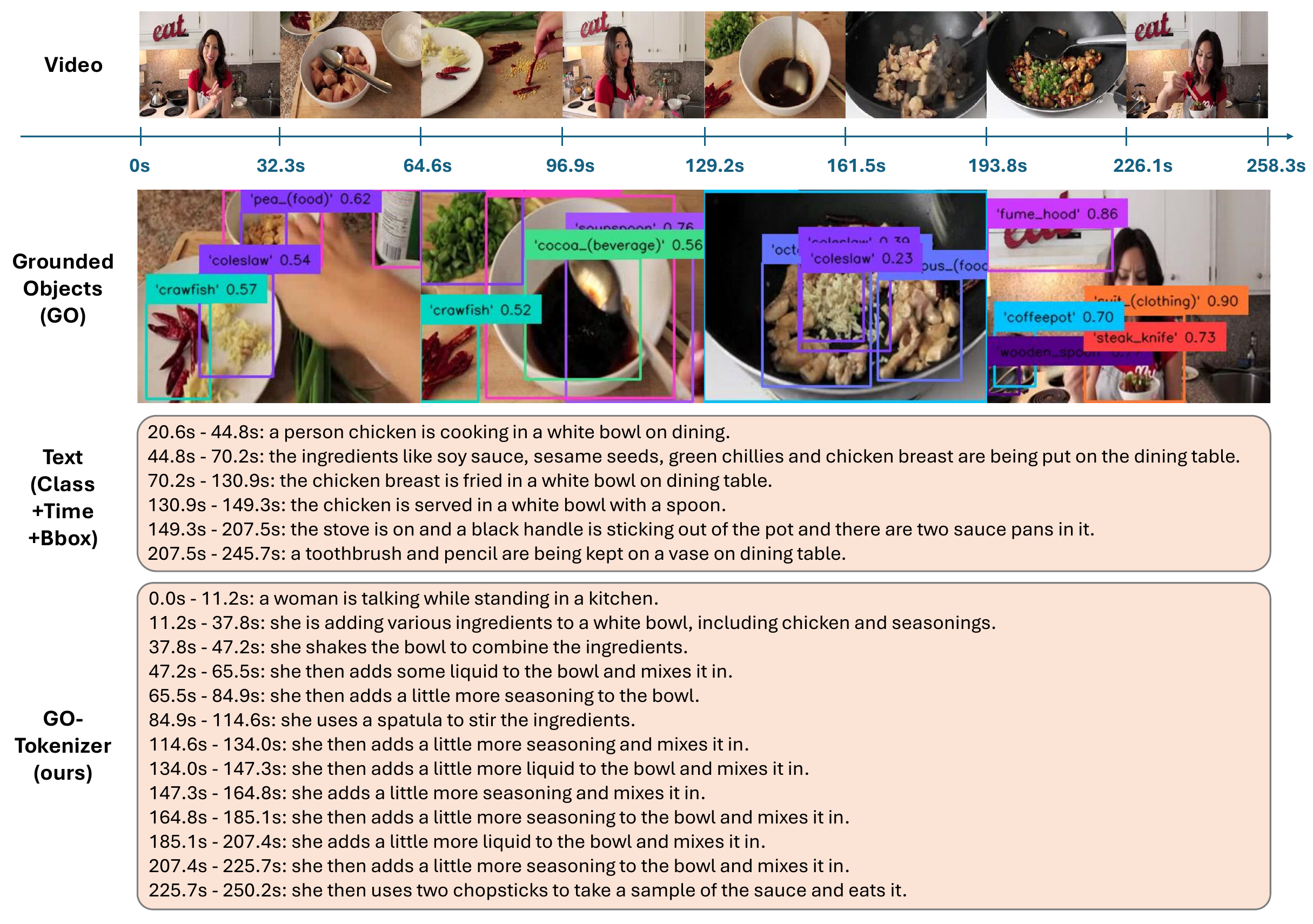}
    \vspace{-22pt}
    \caption{Qualitative results on YouCook-2. The first row shows the frames sampled from the video, the second row shows the objects detected by YOLO-World, and the last two rows show the predicted dense captions from the baseline model with text-based object tokens and with the tokens from GO-Tokenizer, respectively.}
    \label{fig:qualitative}
    \vspace{-13pt}
\end{figure*}

% \vspace{-2pt}
\paragraph{Object detector.}
Two different object detectors are considered in this paper, including Detic~\cite{deticzhou2022detecting} and YOLO-World~\cite{YOLOWorldCheng2024}. As shown in Table~\ref{tab:objec_detector_ablade}, the proposed GO-Tokenizer consistently improves over the text prompt approaches (baselines in section~\ref{sec:baseline}) and the vanilla Video-LLM, regardless of what object detector is being used.

\begin{figure}[t!]
    \centering
    \begin{minipage}[h]{0.48\linewidth}
        \centering        \includegraphics[width=\linewidth,height=0.75\linewidth]{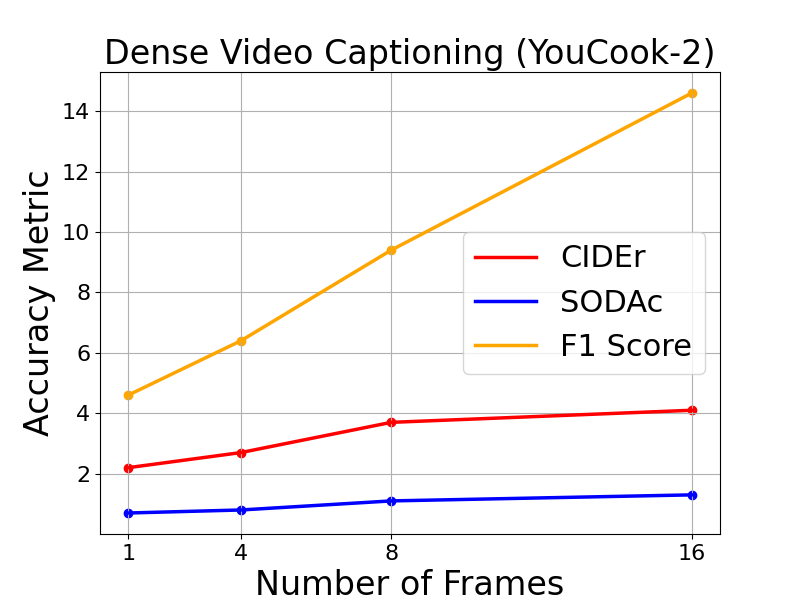}
        \vspace{-8pt}
        \captionof{figure}{Ablation on the number of frames per video (i.e. $F$). Results of GO-TimeChat on YouCook-2 with 1 object per frame using Detic at inference.}
        \label{fig:num_frames}
        \vspace{-5pt}
    \end{minipage}
    \hfill
    \begin{minipage}[h]{0.48\linewidth}
        \centering
    \includegraphics[width=\linewidth,height=0.75\linewidth]{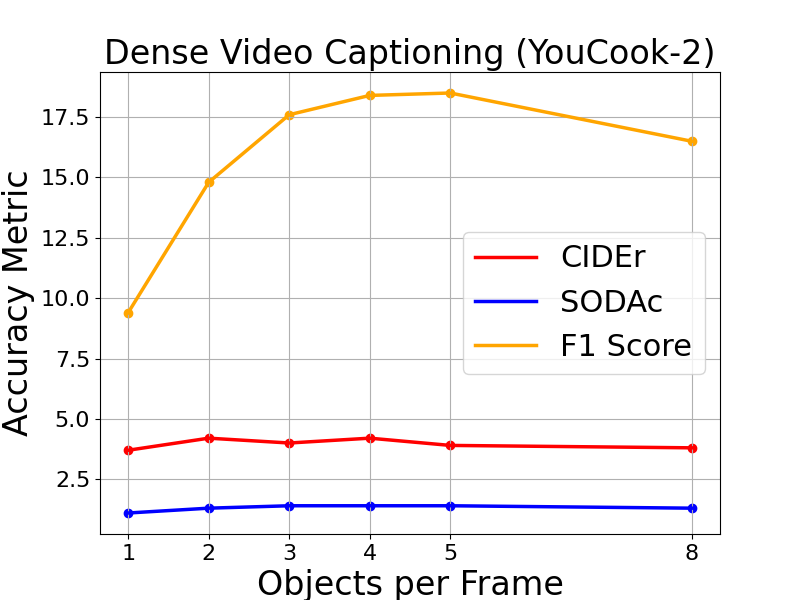}
    \vspace{-8pt}
    \captionof{figure}{Ablation on the number of objects per video frame (i.e. $k$). Results of GO-TimeChat on YouCook-2 with $F=8$ using Detic at inference.}
    \label{fig:num_obj}
    \vspace{-5pt}
    \end{minipage}
    \vspace{-6pt}
\end{figure}

% \vspace{-5pt}
\paragraph{Number of frames.}
To investigate the impact of varying the number of frames containing object information (i.e., $F$), we perform ablation experiments with GO-TimeChat on YouCook-2~\cite{Youcook2ZhXuCoAAAI18}, as shown in Figure~\ref{fig:num_frames}. For these experiments, we keep the number of objects per frame ($k$) fixed to one. The trends in Figure~\ref{fig:num_frames} clearly indicate that increasing the value of $F$ leads to improvements across all the metrics for dense video captioning. In addition, thanks to the compact representation of GO-Tokenizer, scaling to a larger number of frames can be easily achieved.

% \vspace{-5pt}
\paragraph{Number of objects.}
We conduct additional ablation experiments on YouCook-2~\cite{Youcook2ZhXuCoAAAI18} with GO-TimeChat to understand the impact of varying the number of objects per frame ($k$) while keeping the frame count ($F$) constant at 8.
As illustrated in Figure~\ref{fig:num_obj}, we observe that increasing the object count per frame initially improves the accuracy metrics. However, beyond a certain threshold, the metrics either plateau or begin to decline. The best performance is achieved when there are 4 to 5 objects in the frame. We hypothesize that this drop in accuracy is due to a large number of input prompt tokens, which makes it more challenging for the model to process effectively. Additionally, with more objects, the confidence of the object detector also decreases, which could negatively impact the scores as well. 

\vspace{2pt}
\subsection{Qualitative Result}
Figure \ref{fig:qualitative} presents a visualization of zero-shot dense captioning results on YouCook-2 associated with different ways of utilizing GO information. The example shows an instructional video for cooking stir-fried chicken. As shown in the second row, the class names detected by the YOLO-World detector~\cite{YOLOWorldCheng2024} are not always correct (for example, the crawfish is incorrect). This can be due to the vocabulary change between the object detector and the target video dataset, which can introduce a lot of noise to the LLM input. However, GO-Tokenizer does not rely on the decoded class labels and can tolerate minor bounding box drift by its design. Therefore, it generates captions with better quality compared to the baseline models with text prompts. 

%% file: 5_conclusion.tex
\section{Conclusions}
\vspace{-5.5pt}
This work focuses on time-sensitive video understanding (TSV), which requires discerning not only the type of event but also its temporal context, making fine-grained scene comprehension essential for accurate interpretation. We hypothesize that incorporating grounded object (GO) information can enhance TSV tasks and validate this by injecting textual descriptions of objects during inference. However, adding such descriptions for every object across all frames is inefficient and susceptible to prediction noise. To overcome these limitations, we propose GO-Tokenizer, which encodes GO information at a more abstract level. The resulting Video-LLM, integrated with GO-Tokenizer, is referred to as GO-Video and is fine-tuned on the target dataset using low-rank adaptation. Experiments demonstrate that GO-Tokenizer can be applied to various state-of-the-art Video-LLMs (e.g., LITA and TimeChat), with GO-Video outperforming vanilla Video-LLMs on ActivityNet-RTL and YouCook-2. Extensive ablation studies further reveal that GO-Tokenizer offers a more token-efficient and generalizable approach to encoding object information.

%% file: main.bbl
\begin{thebibliography}{10}\itemsep=-1pt

\bibitem{abu2016youtube}
Sami Abu-El-Haija, Nisarg Kothari, Joonseok Lee, Paul Natsev, George Toderici, Balakrishnan Varadarajan, and Sudheendra Vijayanarasimhan.
\newblock Youtube-8m: A large-scale video classification benchmark.
\newblock {\em arXiv preprint arXiv:1609.08675}, 2016.

\bibitem{llama3modelcard}
AI@Meta.
\newblock Llama 3 model card, 2024.

\bibitem{videollmonlinechen2024}
Joya Chen, Zhaoyang Lv, Shiwei Wu, Kevin~Qinghong Lin, Chenan Song, Difei Gao, Jia-Wei Liu, Ziteng Gao, Dongxing Mao, and Mike~Zheng Shou.
\newblock Videollm-online: Online video large language model for streaming video.
\newblock In {\em CVPR}, 2024.

\bibitem{YOLOWorldCheng2024}
Tianheng Cheng, Lin Song, Yixiao Ge, Wenyu Liu, Xinggang Wang, and Ying Shan.
\newblock Yolo-world: Real-time open-vocabulary object detection.
\newblock In {\em Proc. IEEE Conf. Computer Vision and Pattern Recognition (CVPR)}, 2024.

\bibitem{cheng2024videollama2advancingspatialtemporal}
Zesen Cheng, Sicong Leng, Hang Zhang, Yifei Xin, Xin Li, Guanzheng Chen, Yongxin Zhu, Wenqi Zhang, Ziyang Luo, Deli Zhao, and Lidong Bing.
\newblock Videollama 2: Advancing spatial-temporal modeling and audio understanding in video-llms, 2024.

\bibitem{chiang2023vicuna}
Wei-Lin Chiang, Zhuohan Li, Zi Lin, Ying Sheng, Zhanghao Wu, Hao Zhang, Lianmin Zheng, Siyuan Zhuang, Yonghao Zhuang, Joseph~E Gonzalez, et~al.
\newblock Vicuna: An open-source chatbot impressing gpt-4 with 90\%* chatgpt quality.
\newblock {\em See https://vicuna. lmsys. org (accessed 14 April 2023)}, 2(3):6, 2023.

\bibitem{dosovitskiy2020vit}
Alexey Dosovitskiy, Lucas Beyer, Alexander Kolesnikov, Dirk Weissenborn, Xiaohua Zhai, Thomas Unterthiner, Mostafa Dehghani, Matthias Minderer, Georg Heigold, Sylvain Gelly, Jakob Uszkoreit, and Neil Houlsby.
\newblock An image is worth 16x16 words: Transformers for image recognition at scale.
\newblock {\em ICLR}, 2021.

\bibitem{caba2015activitynet}
Bernard~Ghanem Fabian Caba~Heilbron, Victor~Escorcia and Juan~Carlos Niebles.
\newblock Activitynet: A large-scale video benchmark for human activity understanding.
\newblock In {\em Proceedings of the IEEE Conference on Computer Vision and Pattern Recognition}, pages 961--970, 2015.

\bibitem{gabeur2020multi}
Valentin Gabeur, Chen Sun, Karteek Alahari, and Cordelia Schmid.
\newblock Multi-modal transformer for video retrieval.
\newblock In {\em Computer Vision--ECCV 2020: 16th European Conference, Glasgow, UK, August 23--28, 2020, Proceedings, Part IV 16}, pages 214--229. Springer, 2020.

\bibitem{gao2017talltemporalactivitylocalization}
Jiyang Gao, Chen Sun, Zhenheng Yang, and Ram Nevatia.
\newblock Tall: Temporal activity localization via language query, 2017.

\bibitem{guo2024RGPT}
Qiushan Guo, Shalini~De Mello, Hongxu Yin, Wonmin Byeon, Ka~Chun Cheung, Yizhou Yu, Ping Luo, and Sifei Liu.
\newblock Regiongpt: Towards region understanding vision language model.
\newblock In {\em Proceedings of the IEEE/CVF conference on computer vision and pattern recognition}, 2024.

\bibitem{hu2022lora}
Edward~J Hu, Yelong Shen, Phillip Wallis, Zeyuan Allen-Zhu, Yuanzhi Li, Shean Wang, Lu Wang, and Weizhu Chen.
\newblock Lo{RA}: Low-rank adaptation of large language models.
\newblock In {\em International Conference on Learning Representations}, 2022.

\bibitem{Hua2024V2XumLLMCV}
Hang Hua, Yunlong Tang, Chenliang Xu, and Jiebo Luo.
\newblock V2xum-llm: Cross-modal video summarization with temporal prompt instruction tuning.
\newblock {\em ArXiv}, abs/2404.12353, 2024.

\bibitem{litahuang2024}
De-An Huang, Shijia Liao, Subhashree Radhakrishnan, Hongxu Yin, Pavlo Molchanov, Zhiding Yu, and Jan Kautz.
\newblock {LITA}: Language instructed temporal-localization assistant.
\newblock In {\em ECCV}, 2024.

\bibitem{huang2024lita}
De-An Huang, Shijia Liao, Subhashree Radhakrishnan, Hongxu Yin, Pavlo Molchanov, Zhiding Yu, and Jan Kautz.
\newblock Lita: Language instructed temporal-localization assistant.
\newblock In {\em ECCV}, 2024.

\bibitem{jain2024vcoder}
Jitesh Jain, Jianwei Yang, and Humphrey Shi.
\newblock {VCoder: Versatile Vision Encoders for Multimodal Large Language Models}.
\newblock In {\em CVPR}, 2024.

\bibitem{kay2017kineticshumanactionvideo}
Will Kay, Joao Carreira, Karen Simonyan, Brian Zhang, Chloe Hillier, Sudheendra Vijayanarasimhan, Fabio Viola, Tim Green, Trevor Back, Paul Natsev, Mustafa Suleyman, and Andrew Zisserman.
\newblock The kinetics human action video dataset, 2017.

\bibitem{activityNetCaptionskrishna2017dense}
Ranjay Krishna, Kenji Hata, Frederic Ren, Li Fei-Fei, and Juan~Carlos Niebles.
\newblock Dense-captioning events in videos.
\newblock In {\em International Conference on Computer Vision (ICCV)}, 2017.

\bibitem{kuehne2011hmdb}
Hildegard Kuehne, Hueihan Jhuang, Est{\'\i}baliz Garrote, Tomaso Poggio, and Thomas Serre.
\newblock Hmdb: a large video database for human motion recognition.
\newblock In {\em 2011 International conference on computer vision}, pages 2556--2563. IEEE, 2011.

\bibitem{QVHighlights2021}
Jie Lei, Tamara~L. Berg, and Mohit Bansal.
\newblock Qvhighlights: Detecting moments and highlights in videos via natural language queries.
\newblock In {\em NeurIPS}, 2021.

\bibitem{li2024videochatchatcentricvideounderstanding}
KunChang Li, Yinan He, Yi Wang, Yizhuo Li, Wenhai Wang, Ping Luo, Yali Wang, Limin Wang, and Yu Qiao.
\newblock Videochat: Chat-centric video understanding, 2024.

\bibitem{lin2024videollavalearningunitedvisual}
Bin Lin, Yang Ye, Bin Zhu, Jiaxi Cui, Munan Ning, Peng Jin, and Li Yuan.
\newblock Video-llava: Learning united visual representation by alignment before projection, 2024.

\bibitem{liu2023llava}
Haotian Liu, Chunyuan Li, Qingyang Wu, and Yong~Jae Lee.
\newblock Visual instruction tuning.
\newblock In {\em NeurIPS}, 2023.

\bibitem{luo2023valleyvideoassistantlarge}
Ruipu Luo, Ziwang Zhao, Min Yang, Junwei Dong, Da Li, Pengcheng Lu, Tao Wang, Linmei Hu, Minghui Qiu, and Zhongyu Wei.
\newblock Valley: Video assistant with large language model enhanced ability, 2023.

\bibitem{luo2023valley}
Ruipu Luo, Ziwang Zhao, Min Yang, Junwei Dong, Minghui Qiu, Pengcheng Lu, Tao Wang, and Zhongyu Wei.
\newblock Valley: Video assistant with large language model enhanced ability, 2023.

\bibitem{10350951}
Kaijing Ma, Xianghao Zang, Zerun Feng, Han Fang, Chao Ban, Yuhan Wei, Zhongjiang He, Yongxiang Li, and Hao Sun.
\newblock Llavilo: Boosting video moment retrieval via adapter-based multimodal modeling.
\newblock In {\em 2023 IEEE/CVF International Conference on Computer Vision Workshops (ICCVW)}, pages 2790--2795, 2023.

\bibitem{maaz2024videochatgptdetailedvideounderstanding}
Muhammad Maaz, Hanoona Rasheed, Salman Khan, and Fahad~Shahbaz Khan.
\newblock Video-chatgpt: Towards detailed video understanding via large vision and language models, 2024.

\bibitem{Qian2024MomentorAV}
Long Qian, Juncheng Li, Yu Wu, Yaobo Ye, Hao Fei, Tat-Seng Chua, Yueting Zhuang, and Siliang Tang.
\newblock Momentor: Advancing video large language model with fine-grained temporal reasoning.
\newblock {\em ArXiv}, abs/2402.11435, 2024.

\bibitem{qiu2021condensing}
Zhaofan Qiu, Ting Yao, Yan Shu, Chong-Wah Ngo, and Tao Mei.
\newblock Condensing a sequence to one informative frame for video recognition.
\newblock In {\em Proceedings of the IEEE/CVF International Conference on Computer Vision}, pages 16311--16320, 2021.

\bibitem{ClipRadford2021LearningTV}
Alec Radford, Jong~Wook Kim, Chris Hallacy, A. Ramesh, Gabriel Goh, Sandhini Agarwal, Girish Sastry, Amanda Askell, Pamela Mishkin, Jack Clark, Gretchen Krueger, and Ilya Sutskever.
\newblock Learning transferable visual models from natural language supervision.
\newblock In {\em ICML}, 2021.

\bibitem{TimeChatRen2023}
Shuhuai Ren, Linli Yao, Shicheng Li, Xu Sun, and Lu Hou.
\newblock Timechat: A time-sensitive multimodal large language model for long video understanding.
\newblock In {\em CVPR}, 2024.

\bibitem{sharghi2017queryfocusedvideosummarizationdataset}
Aidean Sharghi, Jacob~S. Laurel, and Boqing Gong.
\newblock Query-focused video summarization: Dataset, evaluation, and a memory network based approach, 2017.

\bibitem{Song_2015_CVPR}
Yale Song, Jordi Vallmitjana, Amanda Stent, and Alejandro Jaimes.
\newblock Tvsum: Summarizing web videos using titles.
\newblock In {\em Proceedings of the IEEE Conference on Computer Vision and Pattern Recognition (CVPR)}, June 2015.

\bibitem{tang2024videounderstandingLL}
Yunlong Tang, Jing Bi, Siting Xu, Luchuan Song, Susan Liang, Teng Wang, Daoan Zhang, Jie An, Jingyang Lin, Rongyi Zhu, Ali Vosoughi, Chao Huang, Zeliang Zhang, Pinxin Liu, Mingqian Feng, Feng Zheng, Jianguo Zhang, Ping Luo, Jiebo Luo, and Chenliang Xu.
\newblock Video understanding with large language models: A survey, 2024.

\bibitem{touvron2023llamaopenefficientfoundation}
Hugo Touvron, Thibaut Lavril, Gautier Izacard, Xavier Martinet, Marie-Anne Lachaux, Timothée Lacroix, Baptiste Rozière, Naman Goyal, Eric Hambro, Faisal Azhar, Aurelien Rodriguez, Armand Joulin, Edouard Grave, and Guillaume Lample.
\newblock Llama: Open and efficient foundation language models, 2023.

\bibitem{Wang2024ElysiumEO}
Hang Wang, Yanjie Wang, Yongjie Ye, Yuxiang Nie, and Can Huang.
\newblock Elysium: Exploring object-level perception in videos via mllm.
\newblock In {\em European Conference on Computer Vision}, 2024.

\bibitem{Wu2023VisualCT}
Chenfei Wu, Sheng-Kai Yin, Weizhen Qi, Xiaodong Wang, Zecheng Tang, and Nan Duan.
\newblock Visual chatgpt: Talking, drawing and editing with visual foundation models.
\newblock {\em ArXiv}, abs/2303.04671, 2023.

\bibitem{wu2024freeva}
Wenhao Wu.
\newblock Freeva: Offline mllm as training-free video assistant, 2024.

\bibitem{xu2016msr}
Jun Xu, Tao Mei, Ting Yao, and Yong Rui.
\newblock Msr-vtt: A large video description dataset for bridging video and language.
\newblock In {\em Proceedings of the IEEE conference on computer vision and pattern recognition}, pages 5288--5296, 2016.

\bibitem{xu2024slowfast}
Mingze Xu, Mingfei Gao, Zhe Gan, Hong-You Chen, Zhengfeng Lai, Haiming Gang, Kai Kang, and Afshin Dehghan.
\newblock Slowfast-llava: A strong training-free baseline for video large language models, 2024.

\bibitem{yan2024visa}
Cilin Yan, Haochen Wang, Shilin Yan, Xiaolong Jiang, Yao Hu, Guoliang Kang, Weidi Xie, and Efstratios Gavves.
\newblock Visa: Reasoning video object segmentation via large language models.
\newblock {\em arXiv preprint arXiv:2407.11325}, 2024.

\bibitem{Yang2024MLLMReIDML}
Shan Yang and Yongfei Zhang.
\newblock Mllmreid: Multimodal large language model-based person re-identification.
\newblock {\em ArXiv}, abs/2401.13201, 2024.

\bibitem{yang2023dawnlmmspreliminaryexplorations}
Zhengyuan Yang, Linjie Li, Kevin Lin, Jianfeng Wang, Chung-Ching Lin, Zicheng Liu, and Lijuan Wang.
\newblock The dawn of lmms: Preliminary explorations with gpt-4v(ision), 2023.

\bibitem{damonlpsg2023videollama}
Hang Zhang, Xin Li, and Lidong Bing.
\newblock Video-llama: An instruction-tuned audio-visual language model for video understanding.
\newblock {\em arXiv preprint arXiv:2306.02858}, 2023.

\bibitem{zhang2024llavanext-video}
Yuanhan Zhang, Bo Li, haotian Liu, Yong~jae Lee, Liangke Gui, Di Fu, Jiashi Feng, Ziwei Liu, and Chunyuan Li.
\newblock Llava-next: A strong zero-shot video understanding model, April 2024.

\bibitem{mgllava}
Xiangyu Zhao, Xiangtai Li, Haodong Duan, Haian Huang, Yining Li, Kai Chen, and Hua Yang.
\newblock Towards semantic equivalence of tokenization in multimodal llm.
\newblock In {\em ICLR}, 2025.

\bibitem{ActivityNetEntitieszhou2019grounded}
Luowei Zhou, Yannis Kalantidis, Xinlei Chen, Jason~J Corso, and Marcus Rohrbach.
\newblock Grounded video description.
\newblock In {\em CVPR}, 2019.

\bibitem{Youcook2ZhXuCoAAAI18}
Luowei Zhou, Chenliang Xu, and Jason~J Corso.
\newblock Towards automatic learning of procedures from web instructional videos.
\newblock In {\em AAAI Conference on Artificial Intelligence}, pages 7590--7598, 2018.

\bibitem{Zhou2024StreamingDV}
Xingyi Zhou, Anurag Arnab, Shyamal Buch, Shen Yan, Austin Myers, Xuehan Xiong, Arsha Nagrani, and Cordelia Schmid.
\newblock Streaming dense video captioning.
\newblock {\em 2024 IEEE/CVF Conference on Computer Vision and Pattern Recognition (CVPR)}, pages 18243--18252, 2024.

\bibitem{deticzhou2022detecting}
Xingyi Zhou, Rohit Girdhar, Armand Joulin, Philipp Kr{\"a}henb{\"u}hl, and Ishan Misra.
\newblock Detecting twenty-thousand classes using image-level supervision.
\newblock In {\em ECCV}, 2022.

\end{thebibliography}
